\documentclass[letterpaper, 10 pt, conference]{ieeeconf}  
\renewcommand{\thesection}{\Roman{section}} 
\renewcommand{\thesubsection}{\thesection.\Roman{subsection}}

\IEEEoverridecommandlockouts                              
\usepackage{graphics} 
\usepackage{graphicx}
\usepackage[sort,compress]{cite}

\title{\LARGE Skin Lesion Analysis toward Melanoma Detection: \\A Challenge at the  International Symposium on Biomedical Imaging\\(ISBI) 2016, hosted by the International Skin Imaging Collaboration (ISIC)
}

\author{David Gutman$^{1\dagger}$, Noel C. F. Codella$^{2\dagger}$, Emre Celebi$^{3}$, Brian Helba$^{4}$, \\Michael Marchetti$^{5}$, Nabin Mishra$^{6}$, Allan Halpern$^{5\ddagger}$    
\thanks{*This work was not supported by NCI U24-CA194362-01. This research was funded in part through the NIH/NCI Cancer Center Support Grant P30 CA008748. }
\thanks{$^{\dagger}$ The first two authors contributed equally to this work.}
\thanks{$^{\ddagger}$ Corresponding author.}
\thanks{$^{1}$Emory University, Atlanta GA, USA
        {\tt\small dgutman at emory.edu}}%
\thanks{$^{2}$IBM T. J. Watson Research Center, Yorktown, NY USA
        {\tt\small nccodell at us.ibm.com}}%
\thanks{$^{3}$Louisiana State University in Shreveport, Shreveport, LA USA
        {\tt\small ecelebi at lsus.edu}}%
\thanks{$^{4}$Kitware, Clifton Park, NY  USA
        {\tt\small brian.helba at kitware.com}}%
\thanks{$^{5}$Memorial Sloan-Kettering Cancer Center, New York, NY  USA
		{\tt\small halperna at mskcc.org}}%
\thanks{$^{6}$ Missouri University of Science and Technology, Rolla, MO  USA
		{\tt\small nkmhd3 at mst.edu}}%
}

\begin{document}

\maketitle
\thispagestyle{empty}
\pagestyle{empty}

\begin{abstract}
In this article, we describe the design and implementation of a publicly accessible dermatology image analysis benchmark challenge. The goal of the challenge is to support research and development of algorithms for automated diagnosis of melanoma, a lethal form of skin cancer, from dermoscopic images. The challenge was divided into sub-challenges for each task involved in image analysis, including lesion segmentation, dermoscopic feature detection within a lesion, and classification of melanoma. Training data included 900 images. A separate test dataset of 379 images was provided to measure resultant performance of systems developed with the training data. Ground truth for both training and test sets was generated by a panel of dermoscopic experts. In total, there were 79 submissions from a group of 38 participants, making this the largest standardized and comparative study for melanoma diagnosis in dermoscopic images to date. While the official challenge duration and ranking of participants has concluded, the datasets remain available for further research and development. 

\end{abstract}

\section{INTRODUCTION}

Skin cancer is a major public health concern, with over 5 million newly diagnosed cases in the United States each year \cite{rate,cancerfacts,10k}. Melanoma is one of the most lethal forms of skin cancer, previously responsible for over 9,000 deaths a year in the United States alone ~\cite{cancerfacts}, and over 10,000 estimated deaths in 2016 ~\cite{10k}. 

\begin{figure}[t]
\centering
\includegraphics[width=3.4in]{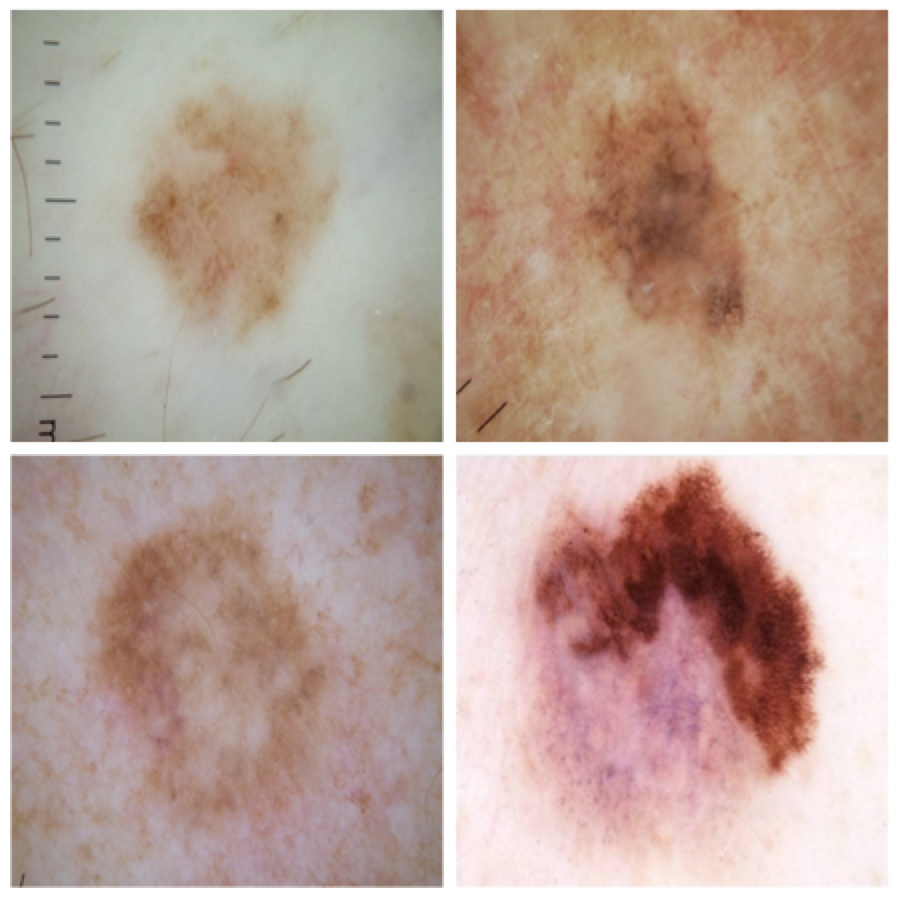}
\caption{Example dermoscopic images of skin lesions. Typical fields-of-view range from 15-30mm at 10X magnification. }
\label{fig:derm}
\end{figure}

As melanoma occurs on the skin surface, it is amenable to detection by simple visual examination. Indeed, most melanomas are first recognized by patients, not physicians ~\cite{patients}. However, unaided visual inspection by expert dermatologists is associated with a diagnostic accuracy of about 60\%, meaning many potential curable melanomas are not detected until more advanced stages \cite{clinicalaccuracy1}. To improve diagnostic performance and reduce melanoma deaths, dermoscopy has been introduced, which is an imaging technique that eliminates the surface reflection of skin, allowing deeper layers to be visually enhanced (Fig. \ref{fig:derm}).  Assuming adequate levels of expertise by the interpreter, dermoscopic imaging has been shown to improve recognition performance over unaided visual inspection by approximately 50\%, resulting in absolute accuracy between 75\%-84\%, with most of the improvement related to an increase in diagnostic sensitivity\cite{MM01,MM02,autodermreview,clinicalaccuracy1,clinicalaccuracy2,kittler}. When clinicians lack expertise, however, no improvement is demonstrated ~\cite{kittler}. Dermoscopic algorithms, such as ``chaos and clues,'' ``3-point checklist,'' ``ABCD rule,'' ``Menzies method,'' ``7-point checklist,'' and ``CASH'' (color, architecture, symmetry, and homogeneity), were developed to facilitate a novice'’s ability to distinguish melanomas from nevi with high diagnostic accuracy ~\cite{clinicalterms1,clinicalterms2,pcp,netmeeting,MM03,MM04,MM05,MM06,MM07,MM08}. However, recent research suggests that many clinicians rely simply on experience and the ``ugly duckling'' sign, which refers to an outlier lesion that is unusual in comparison to other lesions seen on the same patient ~\cite{clinicalprocess}. 


As inexpensive consumer dermatoscope attachments for smart phones are beginning to reach the market ~\cite{molescope}, the opportunity for automated dermoscopic assessment algorithms to positively influence patient care increases.  Given the potential for an influx of images, as well as a growing shortage of dermatologists ~\cite{shortage}, automated tools to assist in triage, screening, and evaluation will become essential. As a result, community interest has grown, as many centers have begun their own research efforts on automated analysis \cite{ph2,codellamlmi,miccaiderm1,miccaiderm2,miccaiderm3,codellaibm,softwaretool,stoecker,rahil,nabin,review}. While initial attempts have been made to create public archives of images to support research and development on automated algorithms for dermoscopic image assessment \cite{ph2dataset}, a large-scale, centralized, coordinated, and comparative effort across institutions has yet to be implemented.


The International Skin Imaging Collaboration (ISIC) is an international effort to improve melanoma diagnosis ~\cite{isic}, which has recently begun efforts to aggregate a publicly accessible dataset of dermoscopy images. This challenge leveraged a database of dermoscopic skin images from the ISIC Data Archive\footnote{https://isic-archive.com/}, which at the time of this publication contains over 10,000 images collected from leading clinical centers internationally, acquired from a variety of devices used at each center. The images are screened for both privacy and quality assurance. The associated clinical metadata has been vetted by recognized melanoma experts. Broad and international participation in image contribution ensures that the dataset contains a representative clinically relevant sample.

The overarching goal of this challenge was to provide a  ``snapshot'' from the ISIC Archive to support development of automated melanoma diagnosis algorithms from dermoscopic images. The challenge was divided into 3 parts corresponding to each stage of lesion analysis: lesion segmentation, lesion dermoscopic feature detection, and lesion classification.  In the following sections, the provided datasets and evaluation metrics used for the challenge, the participation rate, and the top achieved performance levels are described.

\section{CHALLENGE TASKS \& DATASET}

The challenge consisted of 3 tasks: lesion segmentation, dermoscopic feature detection, and disease classification. The second and third components further consisted of two variants, yielding 5 active task parts that teams could participate in. The following discusses the tasks and the supplied training data for each:

\subsection{ Part 1: Lesion Segmentation Task } 
\begin{figure}[t]
\centering
\includegraphics[width=3.4in]{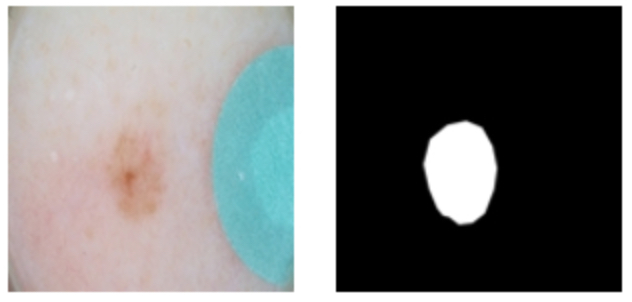}
\caption{Example lesion segmentation. Left: original dermoscopic image. Right: binary segmentation mask.  }
\label{fig:part1}
\end{figure}

Participants were asked to submit automated predictions of lesion segmentations from dermoscopic images in the form of binary masks (Fig. \ref{fig:part1}). Lesion segmentation training data included the original image, paired with the expert manual tracing of the lesion boundaries in the form of a binary mask, where pixel values of 255 are considered inside the area of the lesion, and pixel values of 0 are outside. 900 images and associated ground truth data were supplied for training. Another set of 379 images were provided as a test set from which to evaluate participants.

\subsection{ Part 2: Dermoscopic Feature Classification Task} 
\begin{figure}[t]
\centering
\includegraphics[width=3.4in]{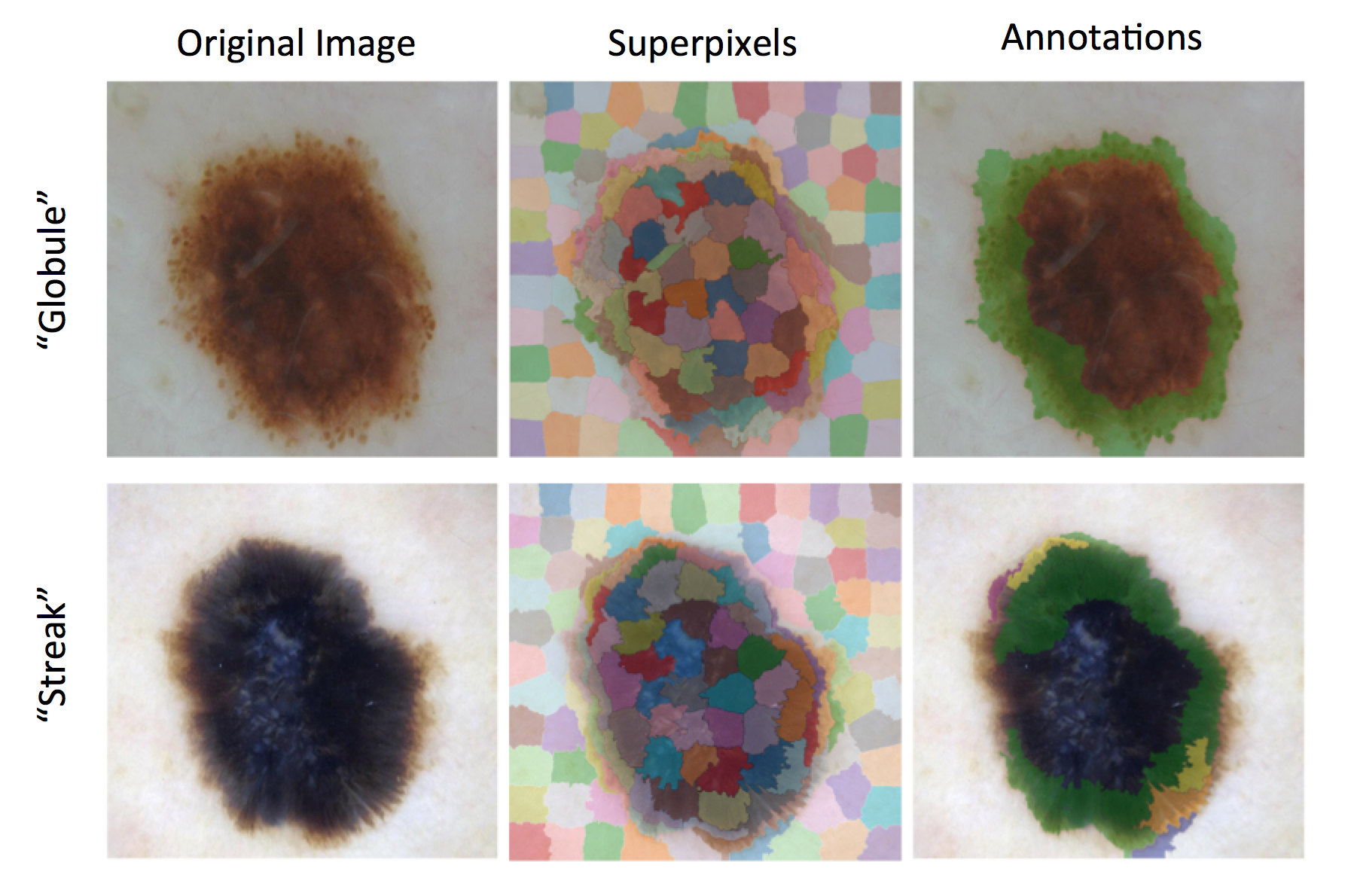}
\caption{Example lesion dermoscopic pattern annotations. Left column: original images. Center column: extracted SLIC superpixels. Right column: Positive superpixel annotations highlighted, overlayed over original image. Multiple colors correspond to multiple human annotators. Top row: example for ``Globule'' annotation label. Bottom row: example for ``Streak'' annotation label.  }
\label{fig:part2}
\end{figure}

Participants were asked to automatically detect two clinically defined dermoscopic features, "globules" and "streaks" ~\cite{clinicalterms1,clinicalterms2}. Pattern detection involved both localization and classification (Fig. \ref{fig:part2}). To reduce the variability and dimensionality of spatial feature annotations, the lesion images were subdivided into superpixels using the SLIC algorithm \cite{slic}. Lesion dermoscopic feature data included the original lesion image and a corresponding superpixel mask, paired with superpixel-wise expert annotations of the presence and absence of the "globules" and "streaks" dermoscopic features. 807 images were provided for training data, and 335 were held-out as a test dataset.

\subsection{ Part 2B: Dermoscopic Feature Segmentation Task } This part was identical to Part 2, with the exception that predictions were in the form of binary masks for each dermoscopic feature. This additional part was provided to explore and compare a second mechanism of algorithm development and evaluation for the goal of lesion dermoscopic pattern detection.

\subsection{ Part 3: Disease Classification Task} Participants were asked to classify images as either being benign or malignant. Prediction classification scores were normalized into confidence intervals from 0.0 (benign) to 1.0 (malignant). Lesion classification data included the original image, paired with both the gold standard (definitive) malignancy diagnosis, as well as the ground truth lesion segmentation. 900 images and associated ground truth data were supplied for training. Another set of 379 images were provided as a test set from which to evaluate participants. Approximately 30.3\% of the dataset was malignant (273 images in the training set).

\begin{figure}[t]
\centering
\includegraphics[width=3.0in]{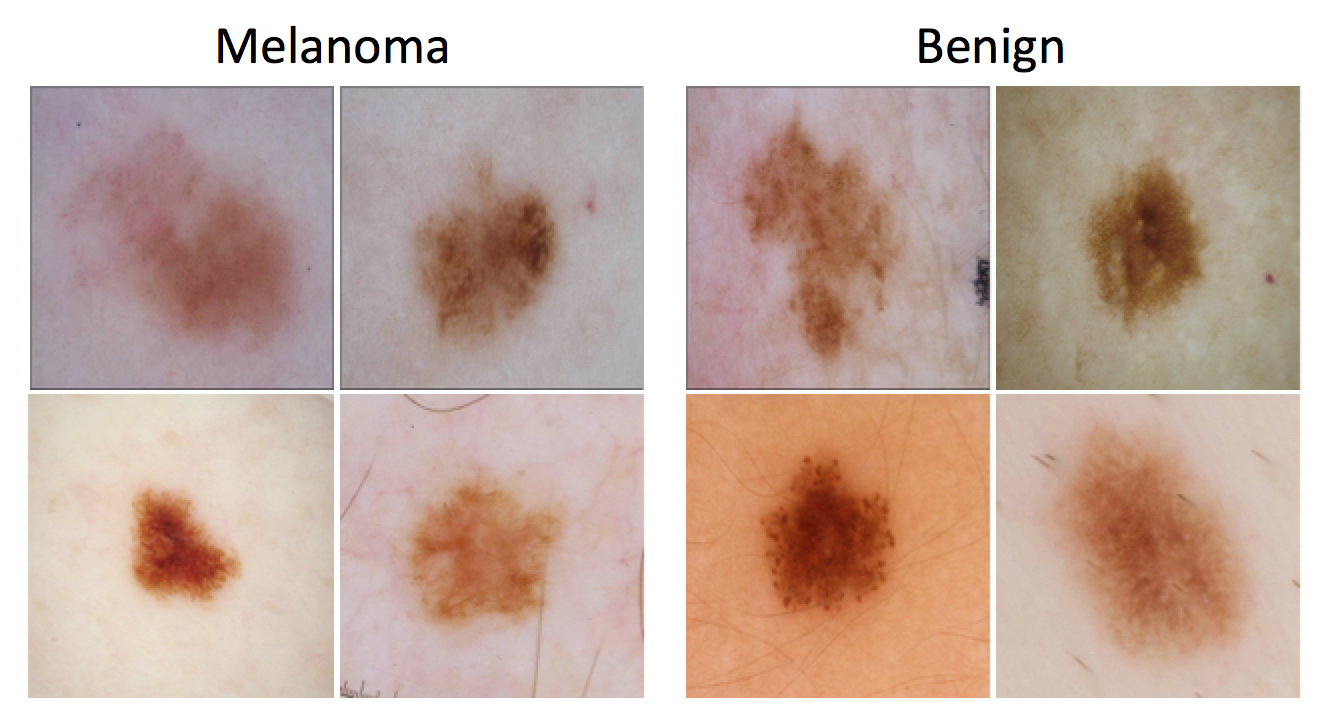}
\caption{Example lesion classification task. Left: 4 example dermoscopic images of melanoma. Right: 4 example dermoscopic images of benign nevi. }
\label{fig:part3}
\end{figure}

\subsection{ Part 3B: Disease Classification Task with Masks} This part was identical to Part 3, with the exception that participants were additionally supplied the ground truth lesion segmentation mask.

\section{EVALUATION CRITERIA}

\subsection{Segmentation Tasks} Submissions were compared  using the following common segmentation metrics: 
\newline

{ \bf Pixel-level accuracy:}
\begin{equation}
AC=\frac{TP+TN}{TP+FP+TN+FN}
\end{equation}

where \emph{TP}, \emph{TN}, \emph{FP}, \emph{FN}, refer to true positive, true negative, false positive, and false negatives, at the pixel level, respectively. Pixel values above 128 were considered positive, and pixel values below were considered negative. 
\newline

{\bf Pixel-level sensitivity:}

\begin{equation}
SE=\frac{TP}{TP+FN}
\end{equation}

{\bf Pixel-level specificity:}

\begin{equation}
SP=\frac{TN}{TN+FP}
\end{equation}

{\bf Dice Coefficient:}
\begin{equation}
DI=\frac{2 \cdot TP}{2 \cdot TP + FN + FP}
\end{equation}

{\bf Jaccard Index:}
\begin{equation}
JA=\frac{TP}{TP + FN + FP}
\end{equation}

Participants were ranked according to the Jaccard. 
\newline

\subsection{Classification Tasks} Submissions were compared using using common classification metrics of accuracy, sensitivity, specificity, as defined in the previous section. However, metrics were measured at the whole image level, rather than the pixel level. Additionally, area under the receiver operating characteristic (ROC) curve (AUC), as well as the specificity at thresholds yielding 95\%  and 98\% sensitivity, were measured. Finally, participants were ranked according to the metric of average precision, evaluated between sensitivity of 0-100\%, which is a common measure of performance in the computer vision community.
\newline

{\bf Area Under Curve:}
\newline
Area under the ROC curve was computed by taking the integral of true positive rate with respect to the false positive rate:
\begin{equation}
AUC=\int_{0}^{1}t_{pr}(f_{pr})\delta f_{pr}
\end{equation}

Where the true positive rate ${t_{pr}}$ is a function of the false positive rate ${f_{pr}}$ along the curve. The ``scikit-learn'' Python package was used for AUC computation. 
\newline

{\bf Average Precision:}
\newline
Assuming dermoscopic images in a dataset are ranked according to normalized machine confidence of melanoma, where the most confident image is at index \emph{k=1}, and \emph{k=n} is the maximum rank that contains all positively labeled instances, the average precision corresponds to the integral under the precision-recall curve within this interval:  

\begin{equation}
AP=\frac{\sum_{k=1}^{n}\left (P(k)\cdot pos(k)  \right )}{n}
\end{equation}

where \emph{k} is an index in the ranked list for evaluation, \emph{pos(k)} is a function that returns 1 if image \emph{k} is a diseased lesion (or 0 otherwise), and \emph{P(k)} is the precision evaluated at index \emph{k}, where precision is defined as the following: 

\begin{equation}
PREC=\frac{TP}{TP + FP}
\end{equation}

The true positive and false positive rate at \emph{k} would be computed by using the machine confidence assigned to image indexed by \emph{k} as the binary threshold between positively and negatively labeled instances. The ``scikit-learn'' Python package was used for computation of average precision.

\section{HOSTING PLATFORM}

The training and test datasets were hosted on the Covalic grand challenge platform, developed at Kitware, Inc.\footnote{http://isic-challenge.net/}, which enabled realtime evaluation of submissions according to defined criteria, automatic ranking of participants based on their submissions, and automatic feedback to participants detailing whether their submissions were properly parsed. Data will continue to be available at this site for the foreseeable future.

\section{RESULTS}

In total, there were 79 submissions from a group of 38 participants (consisting of both individuals and teams). 24 submissions to Part 1, 4 submissions to Part 2, 4 submissions to Part 2B, 25 submissions to Part 3, and 18 submissions to Part 3B. 

Top evaluation results are shown in Tables ~\ref{tab:evalclass} \& ~\ref{tab:evalseg}. Metrics include accuracy (AC), sensitivity (SE), specificity (SP), average precision (AP), area under curve (AUC), specificity at two levels of sensitivity (SP95 \& SP98, at 95\% and 98\% sensitivity, respectively), Dice (DI), and Jaccard (JA). A full listing of results, as well as participants, is available from the challenge website.

\begin{table}[t]
\centering
\caption{Top  evaluation results for each classification task.}
\label{tab:evalclass}
\begin{tabular}{|l|l|l|l|l|l|l|l|} 
\hline
 {\bf Part}       & {\bf AC}   & {\bf SE}  & {\bf SP}  & {\bf AP}    & {\bf AUC} & {\bf SP95} & {\bf SP98} \\ \hline
{\bf 2}  & 0.916 & 0.505 & 0.920 & 0.243 & 0.677 & - & - \\ \hline
{\bf 3}  & 0.855 & 0.507 & 0.941 & 0.637 & 0.804 & 0.227 &  0.095\\ \hline
{\bf 3B} & 0.855 & 0.547 & 0.931 & 0.624 & 0.783 & 0.125 & 0.086 \\ \hline   
\end{tabular}
\end{table}

\begin{table}[t]
\centering
\caption{Top evaluation results for each segmentation task.}
\label{tab:evalseg}
\begin{tabular}{|l|l|l|l|l|l|l|l|l|} 
\hline
 {\bf Part}       & {\bf AC}   & {\bf SE}  & {\bf SP}  & {\bf DI}  & {\bf JA} \\ \hline
{\bf 1}  & 0.953 & 0.910 & 0.965 & 0.910 & 0.843   \\ \hline
{\bf 2B} & 0.962 & 0.396 & 0.968 & 0.128 & 0.070   \\ \hline
\end{tabular}
\end{table}

\section{DISCUSSION}

The results from this public challenge suggest a number of important findings: 1) Performance levels of segmentation methods currently developed appear to be within the range where they would provide utility for annotation of additional data. Further analysis measuring the inter-observer and intra-observer variability of clinical experts will be carried out before a conclusion can be made on whether the automated techniques are statistically equivalent to expert annotation. 2) Results from dermoscopic feature detection appear promising, though further improvements must be made. 3) Disease recognition performance is within the range of that reported in prior literature for expert dermatologists ~\cite{clinicalaccuracy1}. However, further analysis will be done to directly compare automated results on the test dataset to expert blinded dermatologists.


\section{CONCLUSIONS}

Here, we present the design and implementation of a successful public challenge hosted at the 2016 International Symposium on Biomedical Imaging, intended to support the community in the development of automated algorithms for the diagnosis of melanoma from dermoscopic images. A wide variety of independent approaches were submitted and evaluated, yielding the largest standardized and comparative study in this field and topic to date.


\section{ACKNOWLEDGEMENTS}

The authors would like to thank colleagues at Memorial Sloan-Kettering Cancer Center for their work in establishing the ISIC Archive: Ashfaq Marghoob, and Steven Dusza. Additionally, thanks go to co-workers at IBM Research for their support, guidance, and insightful discussions: John R. Smith, Sharath Pankanti, Quoc-Bao Nguyen, and Vaibhava Goel. Finally, thanks go to colleagues at Kitware: Steven Aylward, for his assistance while organizing the challenge, and Max Smolens for technical assistance with the Covalic platform. 

\addtolength{\textheight}{-12cm}   





\end{document}